\documentclass[conference]{IEEEtran}
\IEEEoverridecommandlockouts
\usepackage{cite}
\usepackage{amsmath,amssymb,amsfonts}
\usepackage{algorithm}
\usepackage{algorithmic}
\usepackage{amsmath} 
\usepackage{graphicx}
\usepackage{adjustbox} 
\usepackage{textcomp}
\usepackage{xcolor}
\usepackage{multirow}
\usepackage{url}
\usepackage{subcaption}  
\usepackage{float}       
\usepackage{booktabs}
\usepackage{stfloats}

\begin{document}

\title{HMS-BERT: Hybrid Multi-Task Self-Training for Multilingual and Multi-Label Cyberbullying Detection\\
}

\author{\IEEEauthorblockN{1\textsuperscript{st} Zixin Feng}
\IEEEauthorblockA{\textit{School of AI and Advanced Computing} \\
\textit{XJTLU Entrepreneur College (Taicang)}\\
\textit{Xi’an Jiaotong-Liverpool University}\\
Suzhou 215123, China \\
Zixin.Feng23@student.xjtlu.edu.cn
}
\and
\IEEEauthorblockN{2\textsuperscript{nd} Xinying Cui}
\IEEEauthorblockA{\textit{School of AI and Advanced Computing} \\
\textit{XJTLU Entrepreneur College (Taicang)}\\
\textit{Xi’an Jiaotong-Liverpool University}\\
Suzhou 215123, China \\
Xinying.Cui23@student.xjtlu.edu.cn
}
\and
\IEEEauthorblockN{3\textsuperscript{rd} Yifan Sun}
\IEEEauthorblockA{\textit{School of AI and Advanced Computing} \\
\textit{XJTLU Entrepreneur College (Taicang)}\\
\textit{Xi’an Jiaotong-Liverpool University}\\
Suzhou 215123, China \\
Yifan.Sun2302@student.xjtlu.edu.cn
}
\and
\IEEEauthorblockN{4\textsuperscript{th} Zheng Wei}
\IEEEauthorblockA{\textit{School of AI and Advanced Computing} \\
\textit{XJTLU Entrepreneur College (Taicang)}\\
\textit{Xi’an Jiaotong-Liverpool University}\\
Suzhou 215123, China \\
Zheng.Wei23@student.xjtlu.edu.cn
}
\and
\IEEEauthorblockN{5\textsuperscript{th} Jiachen Yuan}
\IEEEauthorblockA{\textit{School of AI and Advanced Computing} \\
\textit{XJTLU Entrepreneur College (Taicang)}\\
\textit{Xi’an Jiaotong-Liverpool University}\\
Suzhou 215123, China \\
Jiachen.Yuan23@student.xjtlu.edu.cn
}
\and
\IEEEauthorblockN{6\textsuperscript{th} Jiazhen Hu}
\IEEEauthorblockA{\textit{School of AI and Advanced Computing} \\
\textit{XJTLU Entrepreneur College (Taicang)}\\
\textit{Xi’an Jiaotong-Liverpool University}\\
Suzhou 215123, China \\
Jiazhen.Hu24@student.xjtlu.edu.cn
}
\and
\IEEEauthorblockN{7\textsuperscript{th} Ning Xin}
\IEEEauthorblockA{\textit{School of AI and Advanced Computing} \\
\textit{XJTLU Entrepreneur College (Taicang)}\\
\textit{Xi’an Jiaotong-Liverpool University}\\
Suzhou 215123, China \\
Ning.Xin21@student.xjtlu.edu.cn
}
\and
\IEEEauthorblockN{8\textsuperscript{th} Md Maruf Hasan\textsuperscript{*} \\ *Corresponding author}
\IEEEauthorblockA{\textit{School of AI and Advanced Computing} \\
\textit{XJTLU Entrepreneur College (Taicang)}\\
\textit{Xi’an Jiaotong-Liverpool University}\\
Suzhou 215123, China \\
MdMaruf.Hasan@xjtlu.edu.cn
}
}

\date{}

\maketitle

\begin{abstract}
Cyberbullying on social media is inherently multilingual and multi-faceted, where abusive behaviors often overlap across multiple categories. Existing methods are commonly limited by monolingual assumptions or single-task formulations, which restrict their effectiveness in realistic multilingual and multi-label scenarios.  
In this paper, we propose \textbf{HMS-BERT}, a hybrid multi-task self-training framework for multilingual and multi-label cyberbullying detection. Built upon a pretrained multilingual BERT backbone, HMS-BERT integrates contextual representations with handcrafted linguistic features and jointly optimizes a fine-grained multi-label abuse classification task and a three-class main classification task.  
To address labeled data scarcity in low-resource languages, an iterative self-training strategy with confidence-based pseudo-labeling is introduced to facilitate cross-lingual knowledge transfer. Experiments on four public datasets demonstrate that HMS-BERT achieves strong performance, attaining a macro F1-score of up to \textbf{0.9847} on the multi-label task and an accuracy of \textbf{0.6775} on the main classification task. Ablation studies further verify the effectiveness of the proposed components.
\end{abstract}
\begin{IEEEkeywords}
Cyberbullying detection, multilingual text classification, multi-label learning, self-training, transformer-based models
\end{IEEEkeywords}

\section{Introduction}

With the rapid rise of online communication, cyberbullying has become a pervasive and pressing social concern. It involves the transmission of abusive, threatening, or degrading content through digital channels, including text, images, and videos. Compared to traditional bullying, cyberbullying is not limited by time or location. It often occurs anonymously and spreads widely across platforms, causing greater psychological harm to victims~\cite{alhajji2019cyberbullying,vonhumboldt2025words}. Research indicates strong correlations between cyberbullying and increased risks of anxiety, depression, academic failure, and suicidal ideation.

Recent statistics highlight the persistent and widespread impact of cyberbullying, revealing that 55\% of American adolescents have encountered it, with 27\% experiencing it in the last month ~\cite{patchin2024cyberbullying}. Against this backdrop, the growing multilingual nature of user-generated content underscores the need for cyberbullying detection systems that can operate effectively across languages~\cite{saddozai2025multimodal}. Moreover, online abuse often involves overlapping forms such as insults, discrimination, and threats, which pose significant challenges for binary or single-label classifiers~\cite{song2021labelprompt}.

Early research on cyberbullying detection focused primarily on English monolingual data, employing rule-based approaches and traditional machine learning models with hand-made features~\cite{weber2014cyberbullying,betts2016cyberbullying}. With the advent of deep learning, models like Convolutional Neural Networks (CNNs) and Recurrent Neural Networks (RNNs) improved performance by automatically learning contextual representations. Public tools like PerspectiveAPI extended the evaluation of toxicity to large-scale applications, but studies show that they often fail to detect subtle or context-dependent abuse, and their effectiveness is mostly limited to English~\cite{han2020fortifying}. In contrast, offensive language detection in Chinese remains underdeveloped, mainly due to the scarcity of labeled corpora and the lack of publicly available resources~\cite{yang2023sccd}.

Recent progress in multilingual pre-trained language models, such as Multilingual BERT (mBERT), has demonstrated effective cross-lingual transfer capabilities across various NLP tasks~\cite{singh2023mbertgru}. At the same time, resources such as SCCD (Session-based Chinese Cyberbullying Detection Dataset)~\cite{yang2023sccd} have improved the accessibility of annotated Chinese cyberbullying data. However, most existing studies still assume a single-label classification framework, overlooking the overlapping nature of real-world abuse—such as the simultaneous presence of threats, identity attacks, and discrimination. Moreover, despite the growing use of multilingual models, few works have evaluated their effectiveness under multi-label, cross-lingual conditions, particularly in low-resource languages like Chinese.

To address these challenges, we conducted an empirical study on multilingual, multi-label cyberbullying detection using publicly available bilingual datasets comprising Chinese and English online comments. We preprocessed the dataset to ensure consistent multi-label annotations across categories such as threats, discrimination, and identity attacks. Building on the multilingual pre-trained model mBERT, we design \textbf{HMS-BERT}, a hybrid multi-task framework that extends mBERT with feature fusion and self-training mechanisms. This enables the detection of semantically similar yet linguistically diverse abusive expressions, with the aim of improving  performance on low-resource languages like Chinese while maintaining accuracy in English.

\subsection{Contributions}
The main contributions of this work are summarized as follows:
\begin{itemize}
    \item We propose \textbf{HMS-BERT}, a unified multi-task framework that jointly addresses multilingual and multi-label cyberbullying detection across English and Chinese texts.
    \item We introduce an iterative self-training strategy with confidence-based pseudo-labeling to enhance model robustness and cross-lingual generalization under low-resource settings.
    \item We conduct extensive experiments and ablation studies on multiple public datasets to systematically analyze the effectiveness of multi-task learning and self-training for fine-grained cyberbullying detection.
\end{itemize}

In summary, this study provides a comprehensive empirical investigation of cross-lingual, multi-label cyberbullying detection. By applying label normalization and pseudo-labeling strategies on existing public datasets, we evaluate HMS-BERT against multiple baselines. In rigorous experimental settings, we demonstrate its potential in low-resource language scenarios. These findings offer practical implications for the development of multilingual, culturally adaptive cyberbullying detection systems.
\section{Literature Review}

\subsection{Introduction: Machine Learning in Cyberbullying Detection}

Cyberbullying detection techniques have undergone multiple stages of development. Early studies mainly relied on traditional machine learning methods combined with manually designed lexical or syntactic features. Although these approaches achieved certain results, they showed clear limitations in capturing semantic nuances, contextual dependencies, and implicit aggression, especially in cross-domain scenarios.

Aggarwal and Sadhya \cite{mahmud2023cyberbullying} reviewed research on 23 low-resource languages and pointed out challenges in achieving stable generalization under inconsistent labeling systems and limited data scales. Maity et al. \cite{maity2022mtbullygnn} proposed a multitask learning framework combining graph neural networks with sentiment signals to enhance the model’s ability to jointly learn semantics and contextual relations. Musleh et al. \cite{musleh2024machine} focused on Arabic cyberbullying detection and found that the ensemble-based XGBoost model significantly outperformed traditional methods, highlighting the importance of language-specific modeling.

\subsection{The Rise of Pre-trained Language Models (PLMs)}

The introduction of the Transformer architecture marked a major milestone in natural language processing. Vaswani et al. \cite{vaswani2017attention} replaced traditional recurrent structures with a self-attention mechanism, addressing the bottleneck of long-distance dependency modeling and laying the foundation for subsequent large language models. Pires et al. \cite{pires2019multilingualmultilingualbert} analyzed the generalization ability of multilingual BERT (mBERT) and found that the model demonstrated cross-lingual transfer capability even without explicit alignment corpora.

These studies established pre-trained language models as a core component in text analysis and enabled their widespread use in multilingual detection. However, existing pre-trained models still face challenges in task-specific adaptation---while their semantic representations are strong, they often lack specialization for contextual and cultural nuances.

\subsection{Methodological Progress}

Cyberbullying detection methods have gradually transitioned from traditional classifiers to deep neural network models. Mahmud et al. \cite{mahmud2024exhaustive} conducted a systematic comparison of traditional machine learning, deep learning (LSTM/GRU), and pre-trained Transformer models using Bengali and Chittagonian datasets, and found that traditional methods fail to capture contextual and metaphorical aggression.

Kumar et al. \cite{kumar2024bias} further explored Transformer and large language models (LLMs) for detection and data generation, verifying the advantages of the Transformer architecture in feature learning and cross-domain generalization.

Although deep learning has significantly improved detection performance, most studies focus on single-language or domain-specific contexts and lack systematic modeling of multilingual, contextual, and overlapping label characteristics.

\subsection{Multi-label Detection}

Research on multilingual and cross-lingual detection primarily relies on multilingual pre-trained models for knowledge transfer. Singh et al. \cite{singh2023mbertgru} combined mBERT with GRU networks for multilingual hate speech detection, demonstrating strong cross-lingual capability in low-resource settings. Sohn and Lee \cite{sohn2019mcbert4hate} proposed a multi-channel BERT (MC-BERT4HATE) that encodes different languages or translation channels in parallel, improving semantic alignment and robustness to noise.

Future studies should consider integrating mBERT representations with linguistic features to comprehensively model multilingual and multidimensional bullying behaviors.

\subsection{The Multi-faceted Nature of Cyberbullying}

Cyberbullying texts often contain multiple overlapping forms of aggression, making single-label classification inadequate for representing semantic complexity. Song et al. \cite{song2023label} proposed a label-prompt-based multi-label classification framework, in which each potential label is represented by a learnable prompt vector. This design allows the model to leverage the self-attention mechanism to capture inter-label dependencies. Therefore, the model becomes more sensitive to fine-grained and composite aggression categories.

This study is closely related to the present research, demonstrating that incorporating label semantics enhances the model’s ability to detect multidimensional aggressive behaviors. However, the application of multi-label learning in low-resource languages remains limited, and data scarcity and class imbalance continue to restrict its generalization performance.

\subsection{Data Scarcity in Low-Resource Settings}

Labeled data scarcity is a primary challenge in cyberbullying detection, particularly for low-resource languages like Chinese. High-quality annotations are costly to produce, and the subjective, ambiguous nature of offensive language complicates the labeling process. To address this issue, Ghiasi et al. \cite{ghiasi2021multi} proposed MuST, a self-training-based multitask learning framework that training data through pseudo-labeling. This approach improves both feature learning and model stability during multitask optimization.

This method effectively mitigates the issue of limited labeled data and promotes cross-task knowledge sharing. 

\subsection{Research Gap}

Whereas cyberbullying detection has evolved from traditional models to Transformer-based deep models and achieved progress in multilingual, multi-label, and self-training directions, significant gaps remain.

Current studies have not yet integrated these techniques within a unified framework. Existing methods are fragmented and lack a comprehensive model capable of simultaneously addressing the following issues:

\begin{itemize}
    \item Cross-lingual modeling based on mBERT to process multilingual text;
    \item Multi-label learning to capture fine-grained and overlapping bullying categories;
    \item Integration of semantic embeddings with handcrafted linguistic features (e.g., punctuation frequency, sensitive word indicators) to enhance contextual understanding;
    \item Self-training strategies to address label scarcity in low-resource languages.
\end{itemize}

To address these gaps, this study proposes a unified multilingual and multitask detection framework. The framework combines semantic representations with handcrafted linguistic features and incorporates a self-training mechanism. This approach improves model robustness and generalization across multilingual and multi-label settings.

\section{Methodology}

\subsection{Data Preprocessing}
This study employs a unified data preprocessing pipeline to ensure the quality, structure, and consistency of multilingual cyberbullying datasets before model training. As illustrated in Fig.\ref{fig:data-preprocessing-flow}, the preprocessing includes the following essential steps:
\begin{figure}[htbp]
    \centering
    \includegraphics[width=0.50\textwidth]{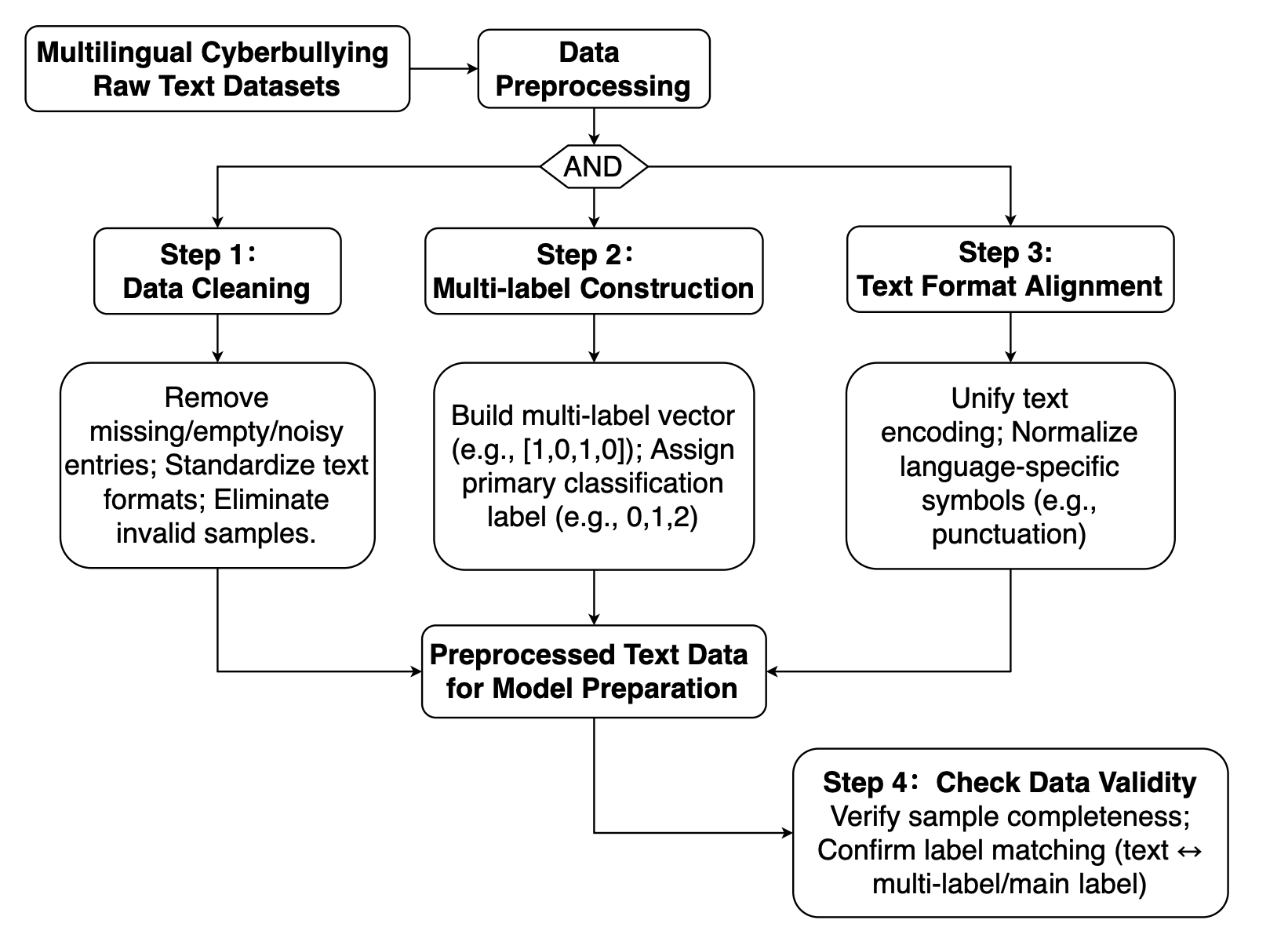}
    \caption{Flowchart of Data Preprocessing}
    \label{fig:data-preprocessing-flow}
\end{figure}

\begin{enumerate}
    \item \textbf{Data cleaning} \\
    Raw datasets from multiple sources are cleaned to remove missing, empty, or noisy entries, standardize text formats (e.g., unified character encoding and capitalization), and eliminate invalid samples.

    \item \textbf{Multi-label construction} \\
    Each cleaned sample is annotated with a multi-label vector (representing different types of bullying) and a primary classification label (representing the overall information intention, e.g., normal, offensive, or hateful).

    \item \textbf{Text format alignment} \\
    The cleaned and labeled texts are normalized in terms of language-specific symbols (e.g., punctuation) and unified into a consistent text encoding format to adapt to subsequent processing.

    \item \textbf{Data validity check} \\
    The processed samples are verified for completeness (e.g., no missing text/label pairs) and validity (e.g., matching between text content and annotation labels) to ensure the availability of preprocessed data.
\end{enumerate}

\subsection{Workflow}
\begin{figure*}
    \centering
    \includegraphics[width=0.7\linewidth]{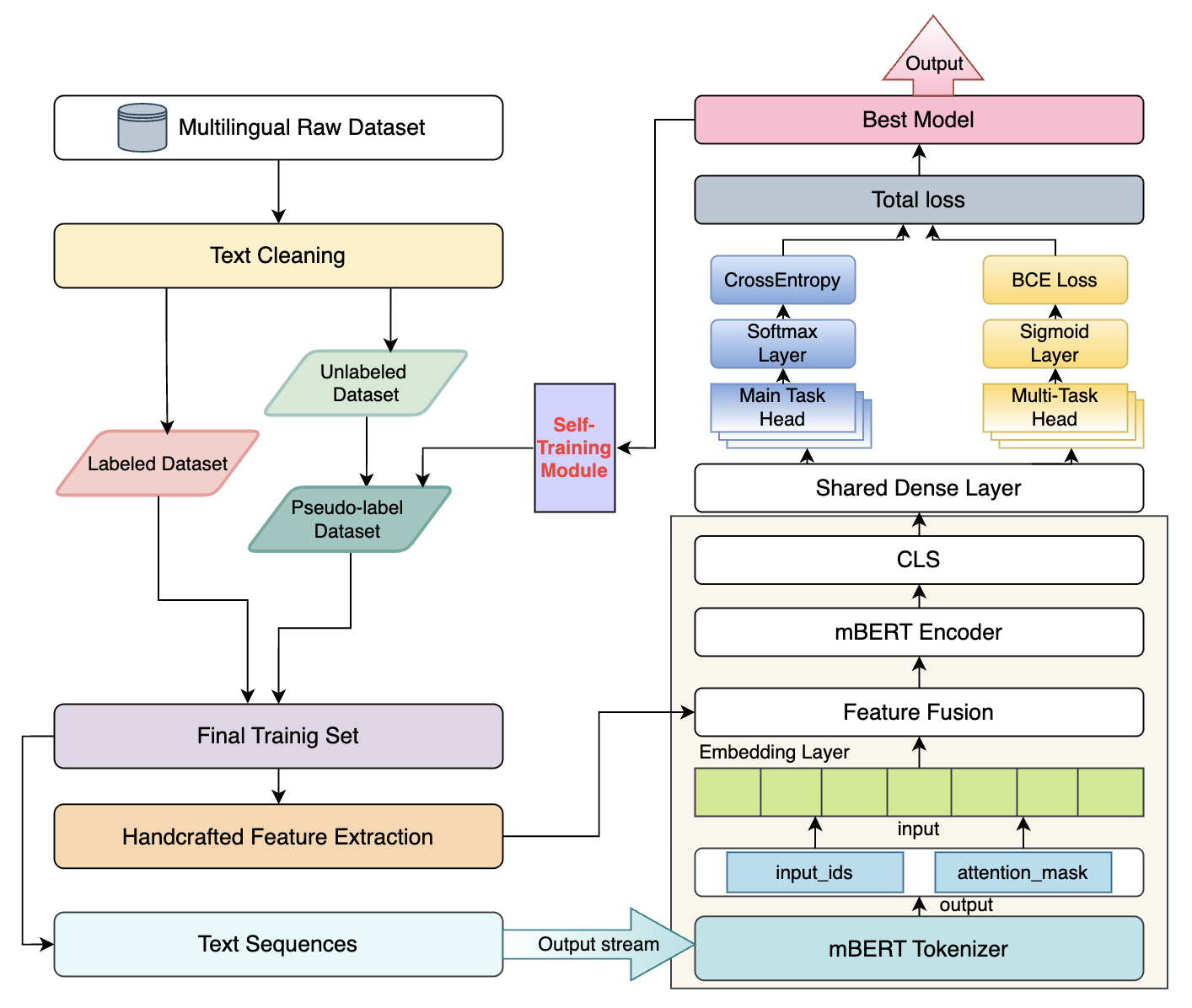} 
    \caption{The workflow of the multilingual input representation, semantic encoding, feature enhancement, and multi-label classification.}
    \label{fig:workflow}
\end{figure*}

To address the challenges of multilingual and multi-label cyberbullying detection, we design a modular framework consisting of four components: input representation, semantic encoding, feature enhancement, and classification. As shown in Fig.~\ref{fig:workflow}, this pipeline enables joint processing of English and Chinese texts, allowing the model to capture co-occurring abuse categories in a unified architecture.~\cite{sohn2019mcbert4hate}.

\subsubsection{\textbf{Input Representation and Feature Fusion}}

After standard preprocessing (e.g., noise removal, token normalization), each input is transformed into two parallel feature streams. The first encodes contextual semantics using the multilingual BERT (mBERT) tokenizer, producing language-agnostic embeddings.~\cite{devlin2019bert}. The second extracts handcrafted lexical features, such as emoji use, punctuation frequency, character length, and binary indicators for sensitive terms. These features help detect both explicit and implicit abuse patterns across languages. The two streams are concatenated after the encoding stage, where the handcrafted feature vector is fused with the mBERT [CLS] representation to form a unified vector for the final classification.

\subsubsection{\textbf{Contextual Embedding with mBERT}}

The fused input is passed through mBERT’s 12-layer Transformer encoder~\cite{devlin2019bert}, which leverages the [CLS] token for sentence-level representation and Multilingual attention to align semantically equivalent expressions. Trained on multilingual corpora, mBERT effectively captures shared semantics across languages. The sentence-level representation is derived from the [CLS] token, which captures both local and global semantics. mBERT’s multilingual capacity allows it to align semantically equivalent expressions across languages through cross-lingual training on multilingual corpora~\cite{sohn2019mcbert4hate}.
\begin{figure*}
    \centering
    \includegraphics[width=0.75\linewidth]{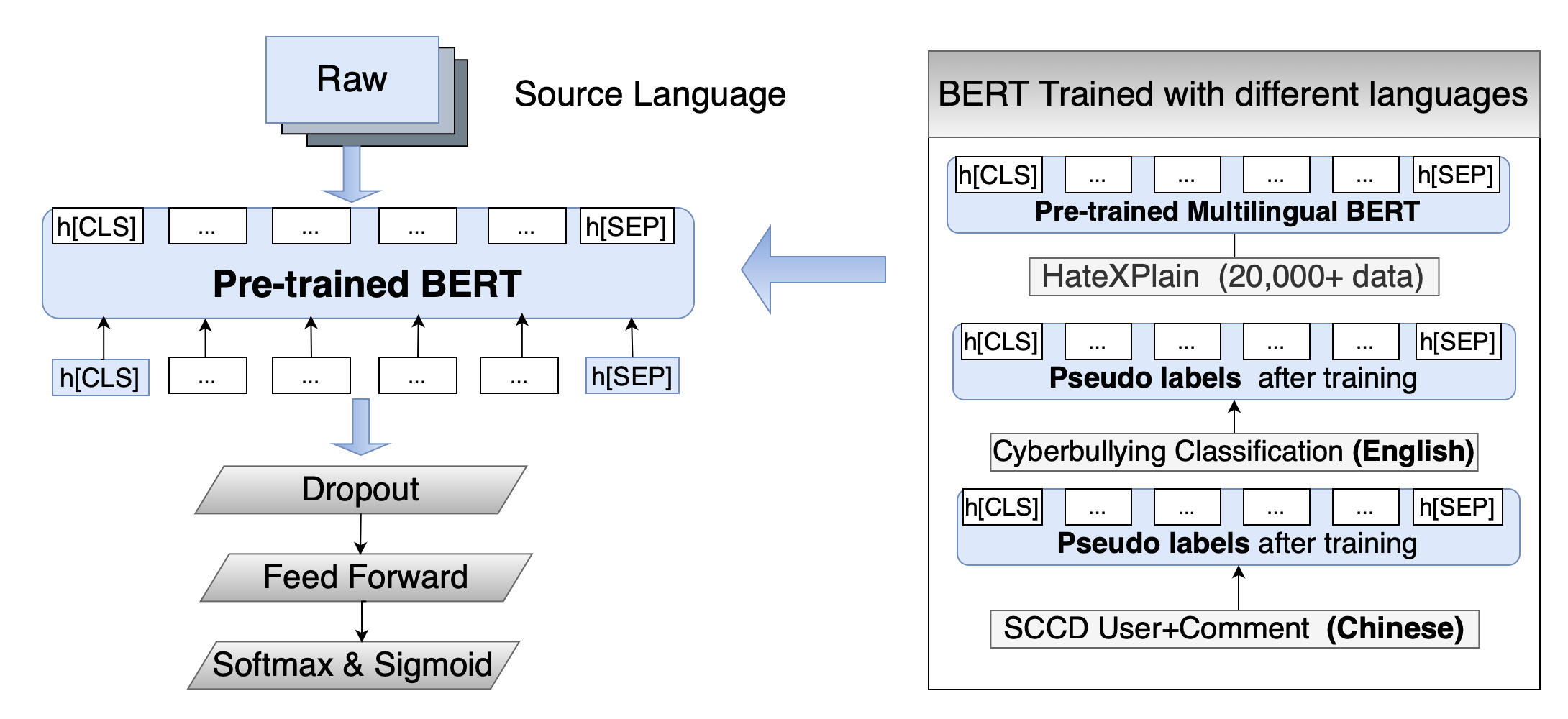} 
    \caption{Step2-Step3: Pre-trained network to Fine-tuned network}
    \label{fig:Pre-trained}
\end{figure*}
\subsubsection{\textbf{Multi-label Classification and Main Task Classification}}

The contextual embeddings are processed through two parallel classification branches. For the multi-label task, a fully connected layer with sigmoid activation is employed to support independent predictions for each label. This allows modeling overlapping abuse types (e.g., identity attack and threat). Concurrently, a three-class main task classification (Hate/Offensive/Normal) is introduced, using a softmax-activated fully connected layer for exclusive category assignment based on the [CLS] token embeddings.

\subsubsection{\textbf{Fine-tuning and Feature Integration}}

To improve robustness, handcrafted features are processed through a dropout layer (dropout rate = 0.3) and a dense layer before integration into the final classification~\cite{singh2023mbertgru}. The model is trained end-to-end using binary cross-entropy for the multi-label task and categorical cross-entropy for the main task. The total loss is the sum of both, optimized using AdamW with a learning rate of $5 \times 10^{-5}$.

\subsubsection{\textbf{Training Strategy and Inference}}

Training is conducted for 3 epochs with a batch size of 16. Learning rate decay and early stopping are applied to avoid overfitting. Inference involves thresholding sigmoid outputs to produce binary label decisions, tuned for balanced precision-recall. The entire pipeline is GPU-accelerated and supports real-time detection in large-scale applications.

\subsection{Baseline Model and Proposed Innovations}
\subsubsection{mBERT-Based Model Architecture}

The model is constructed on top of the \textit{bert-base-multilingual-cased} (mBERT) encoder. mBERT is chosen for its capability to handle cross-lingual text and extract rich semantic features. The architecture consists of three main parts, as illustrated in Fig. \ref{fig:Pre-trained}:

\paragraph{Input Tokenization and Embedding}

Input text \(X\) is tokenized into subwords \(T=\{t_1, t_2, \dots, t_n\}\). mBERT then generates contextual embeddings, outputting a hidden state matrix \(H \in \mathbb{R}^{n \times d}\), where \(H_i\) represents the embedding of subword \(t_i\) and \(d\) is the hidden dimension.

\paragraph{Semantic Representation Extraction}
The embedding of the initial token ([CLS]), which is widely used to capture sequence-level semantics, is extracted as the semantic representation:\[h = H_1 \in \mathbb{R}^{d}.\]
This representation \(h\) is subsequently employed in downstream tasks.  

\paragraph{Dual-Task Prediction Heads}
To support the multi-label subtype prediction task, a sigmoid-activated linear transformation\cite{song2023labelprompt} is applied to project the semantic representation $h$ into $C$ subtype probabilities. The output is given by the classifier function in Eq.~\eqref{eq:multilabel}:
\begin{equation}
\hat{y} = \sigma(W_m h + b_m),
\label{eq:multilabel}
\end{equation}
where $W_m \in \mathbb{R}^{C \times d}$ is the weight matrix, $b_m$ is the bias vector, and $\sigma(\cdot)$ denotes the element-wise sigmoid function. Here, $C$ is the number of bullying subtypes.

For the main tone classification task, a linear transformation followed by a softmax activation \cite{song2023labelprompt} is utilized to project the semantic representation \( h \) into probabilities over \( K \) mutually exclusive tone categories. The final prediction corresponds to the output distribution in Eq.~\eqref{eq:softmax_classification}:
\begin{equation}
\hat{z} = \text{softmax}(W_s h + b_s),
\label{eq:softmax_classification}
\end{equation}
where \( W_s \in \mathbb{R}^{K \times d} \) is the weight matrix, \( b_s \) is the bias vector, and \( \text{softmax}(\cdot) \) denotes the softmax function that transforms the logits into a probability distribution across the \( K \) tone categories (e.g., hate, offensive, normal), ensuring that the prediction is mutually exclusive and corresponds to the category with the highest probability.

\subsubsection{Joint Loss Function}
To enable simultaneous optimization of the model for both multi-label bullying subtype prediction and main tone classification, we introduce a weighted joint loss function that integrates the two task-specific loss components. This design allows the model to learn shared representations beneficial to both tasks while addressing their distinct learning objectives.

For the multi-label subtype classification (with $K$ subtypes), we employ binary cross-entropy (BCE) loss, as defined in Eq.~\eqref{eq:multi_loss}:
\begin{equation}
\mathcal{L}_{\text{multi}} = -\sum_{k=1}^{K} \left[y_k \log \hat{y}_k + (1 - y_k) \log (1 - \hat{y}_k)\right] ,
\label{eq:multi_loss}
\end{equation}
where $y_k \in \{0,1\}$ indicates the presence of the $k$-th bullying subtype, and $\hat{y}_k$ represents the predicted probability.

The categorical cross-entropy (CE) loss for main tone (with $C$ categories) classification is formulated in Eq.~\eqref{eq:main_loss}:
\begin{equation}
\mathcal{L}_{\text{main}} = -\sum_{c=1}^{C} z_c \log \hat{z}_c ,
\label{eq:main_loss}
\end{equation}
where $z_c \in \{0,1\}$ is the one-hot encoded ground-truth label for the $c$-th tone category.

The final joint loss combines these components with a balancing weight $\lambda$, as shown in Eq.~\eqref{eq:joint_loss}:
\begin{equation}
\mathcal{L}_{\text{total}} = \lambda \mathcal{L}_{\text{multi}} + (1 - \lambda) \mathcal{L}_{\text{main}}.
\label{eq:joint_loss}
\end{equation}
We set $\lambda = 0.7$ to emphasize the multi-label task, as it requires more fine-grained discrimination. This weighting scheme ensures balanced optimization across both tasks while prioritizing the more challenging subtype prediction.

\subsubsection{Self-Training Optimization}
To address data scarcity, we integrate a self-training loop that iteratively leverages unlabeled data:  

\begin{enumerate}
\item\textbf{Initialization}: Train the model on labeled data $\mathcal{D}_L$.
    
\item\textbf{Pseudo-Labeling}: For unlabeled data $\mathcal{D}_U$, generate predictions and select high-confidence samples ($\tau > 0.9$) as pseudo-labeled data $\mathcal{D}_{PL}$.
    
\item\textbf{Expanded Training}: Retrain the model on $\mathcal{D}_L \cup \mathcal{D}_{PL}$.
    
\item\textbf{Iteration}: Repeat Steps 2--3 for $T$ iterations until convergence.
\end{enumerate}

During self-training, we implement a confidence decay mechanism: once the model demonstrates stable performance on pseudo-labels (e.g., high consistency in predictions), confidence thresholds are gradually relaxed (e.g., from 0.9 to 0.85). This balances early-stage reliance on high-quality pseudo-labels with late-stage exploration of long-tail examples, ensuring both convergence speed and generalization.

\begin{algorithm}
\caption{Self-training with Pseudo Labels (with Confidence Decay)}
\begin{algorithmic}[1]
\REQUIRE Labeled set $\mathcal{D}_L$, Unlabeled set $\mathcal{D}_U$, Initial confidence threshold $\tau_{\text{init}} = 0.9$, Minimum confidence threshold $\tau_{\text{min}} = 0.85$, Number of iterations $T$, Decay rate $\alpha$ (e.g., $0.02$ per iteration)
\ENSURE Trained model $f$

\STATE Train initial model $f^{(0)}$ on $\mathcal{D}_L$ using joint loss (Eq. 5)
\STATE Initialize $\tau \leftarrow \tau_{\text{init}}$
\FOR{$t = 1$ \TO $T$}
    \FOR{each $x_u \in \mathcal{D}_U$}
        \STATE $\hat{y}_u \leftarrow f^{(t-1)}(x_u)$
        \IF{$\max(\hat{y}_u) > \tau$}
            \STATE $\mathcal{D}_L \leftarrow \mathcal{D}_L \cup \{(x_u, \hat{y}_u)\}$
        \ENDIF
    \ENDFOR
    \STATE Train $f^{(t)}$ on updated $\mathcal{D}_L$ using joint loss (Eq. 5)
    \STATE Update confidence threshold: $\tau \leftarrow \max(\tau_{\text{min}}, \tau - \alpha)$
\ENDFOR
\STATE $f \leftarrow f^{(T)}$
\end{algorithmic}
\end{algorithm}

\section{Experiment}
\subsection{Dataset Overview}
\subsubsection{Dataset Description}

To support multilingual and multi-label cyberbullying detection, we use four real-world datasets covering both English and Chinese. These datasets vary in language, label structure, and usage within our framework. Table~\ref{tab:datasets} summarizes their key properties.
\begin{table}[htbp]
    \centering
    \caption{Summary of datasets used in the study}
    \label{tab:datasets}
    \small
    \begin{tabular}{@{}lcl@{\hspace{3pt}}p{3.5cm}@{}}
        \toprule
        \textbf{Dataset} & \textbf{L} & \textbf{Size} & \textbf{Usage} \\ 
        \midrule
        HateXplain & E & 20K+ & Multi-label + 3-class training \\ 
        \addlinespace[1pt]
        Cyberbullying Classif. & E & 47K+ & Multi-label pseudo-labeling \\ 
        \addlinespace[1pt]
        SCCD\_User & C & 10K+ & Pseudo-labeling (2K eval.) \\ 
        \addlinespace[1pt]
        SCCD\_Comment & C & 10K+ & Pseudo-labeling (2K eval.) \\ 
        \bottomrule
    \end{tabular}
    \vspace{-2mm}
\end{table}

\subsubsection{Dataset Sources and Characteristics}

To support multilingual and multi-label cyberbullying detection, we utilize three real-world datasets selected for their linguistic diversity and annotation structure.
\begin{figure*}[htbp]
    \centering  
    \setlength{\belowcaptionskip}{5pt} 
    
    \begin{subfigure}[b]{0.48\textwidth}
        \centering
        \includegraphics[width=\linewidth,height=5cm,keepaspectratio]{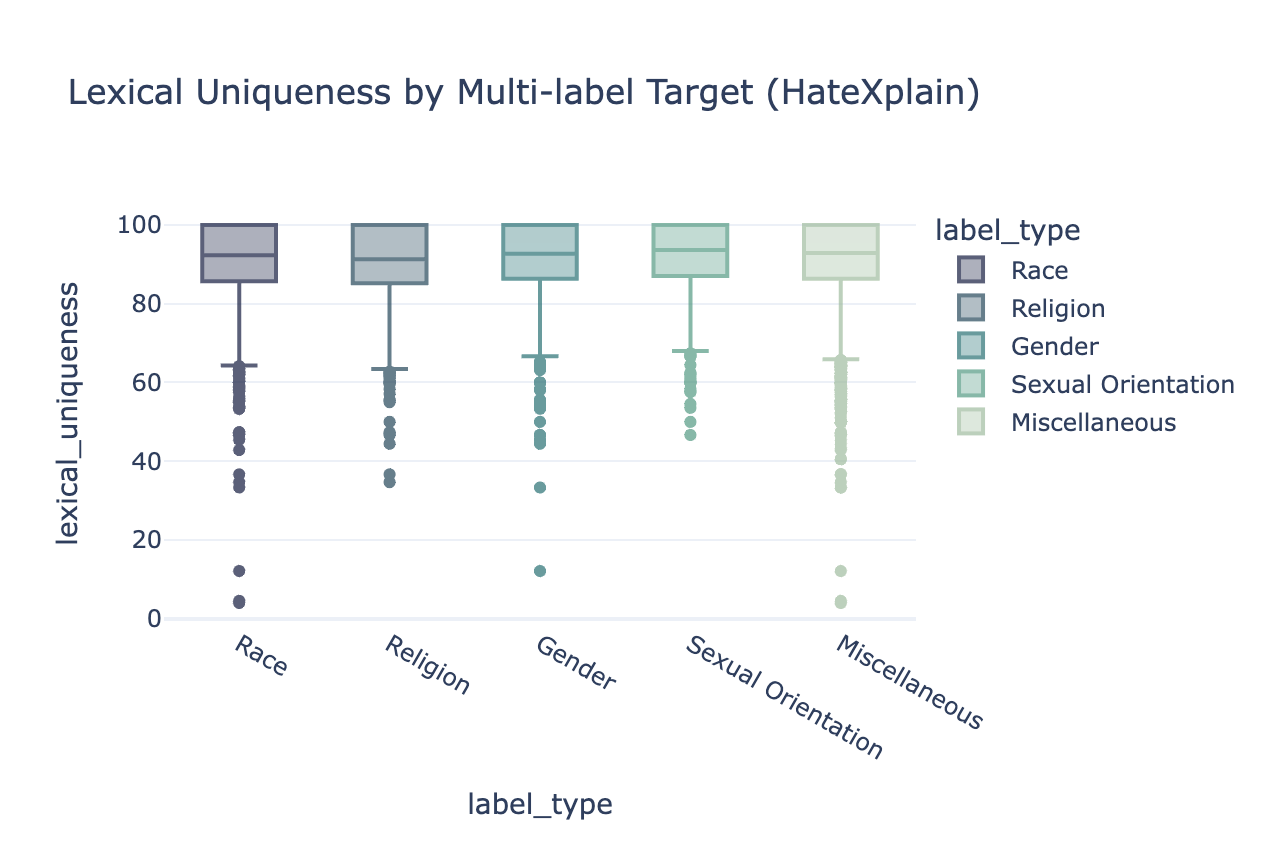}
        \caption{Lexical Uniqueness}
        \label{subfig:lexical}
    \end{subfigure}
    \hspace{2pt}
    \begin{subfigure}[b]{0.48\textwidth}
        \centering
        \includegraphics[width=\linewidth,height=5cm,keepaspectratio]{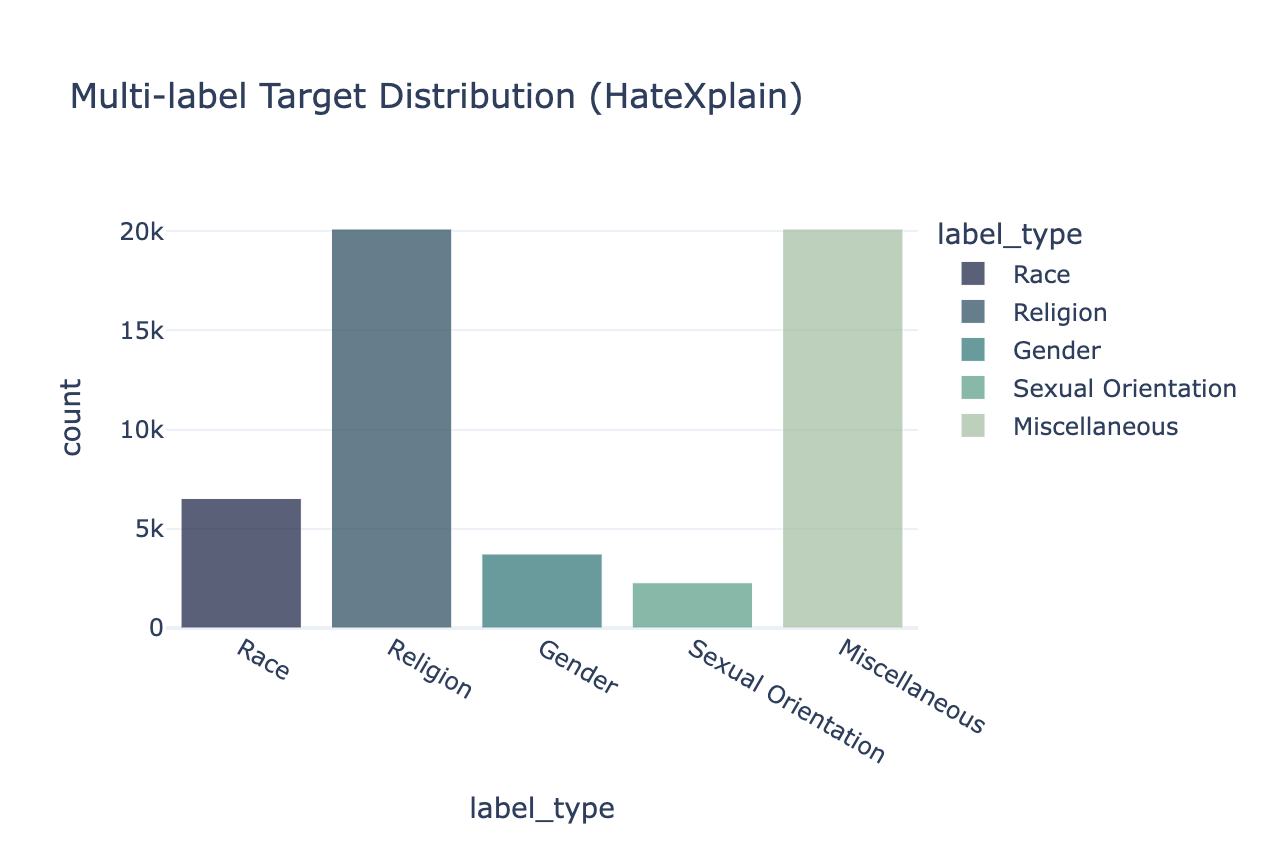}
        \caption{Label Distribution}
        \label{subfig:distribution}
    \end{subfigure}
    
    \vspace{3pt} 
    \begin{subfigure}[b]{0.48\textwidth}
        \centering
        \includegraphics[width=\linewidth,height=5cm,keepaspectratio]{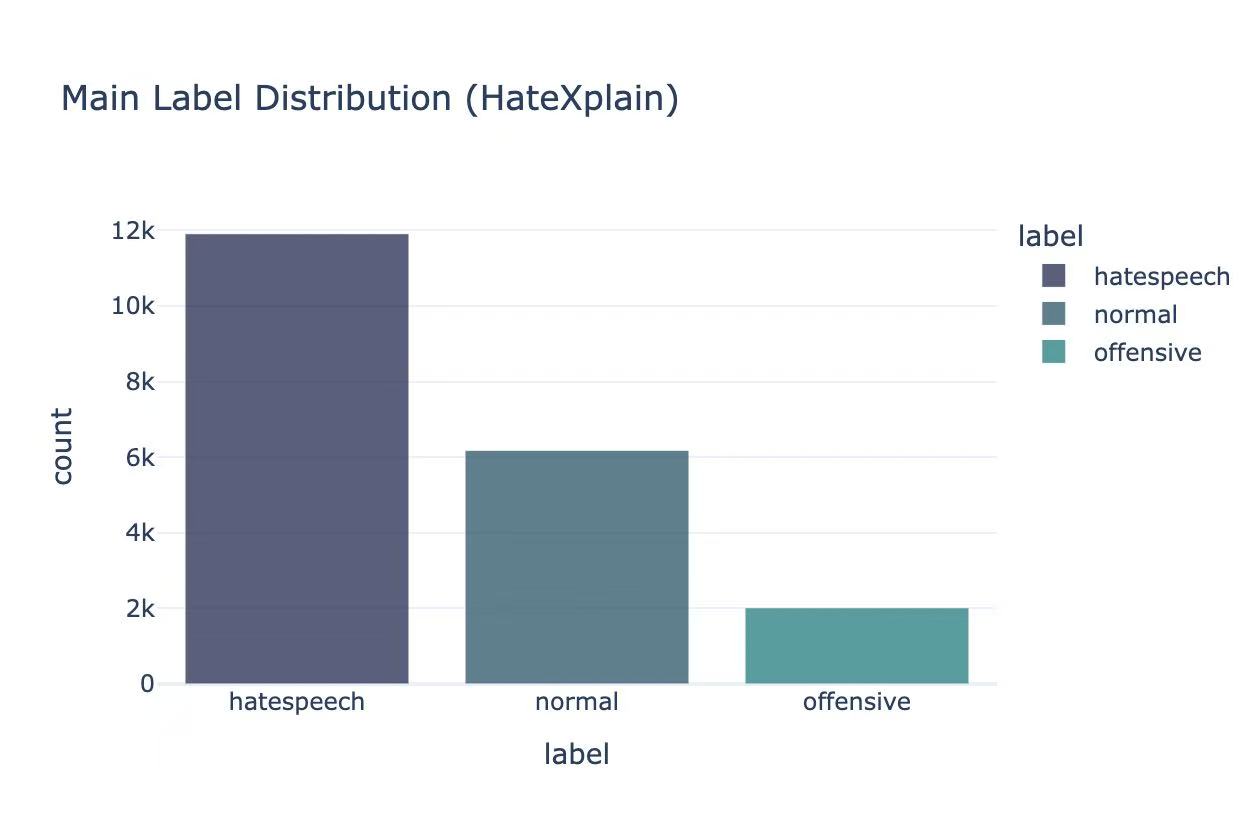}
        \caption{Main Label}
        \label{subfig:主标签}
    \end{subfigure}
    \hspace{2pt}
    \begin{subfigure}[b]{0.48\textwidth}
        \centering
        \includegraphics[width=\linewidth,height=5cm,keepaspectratio]{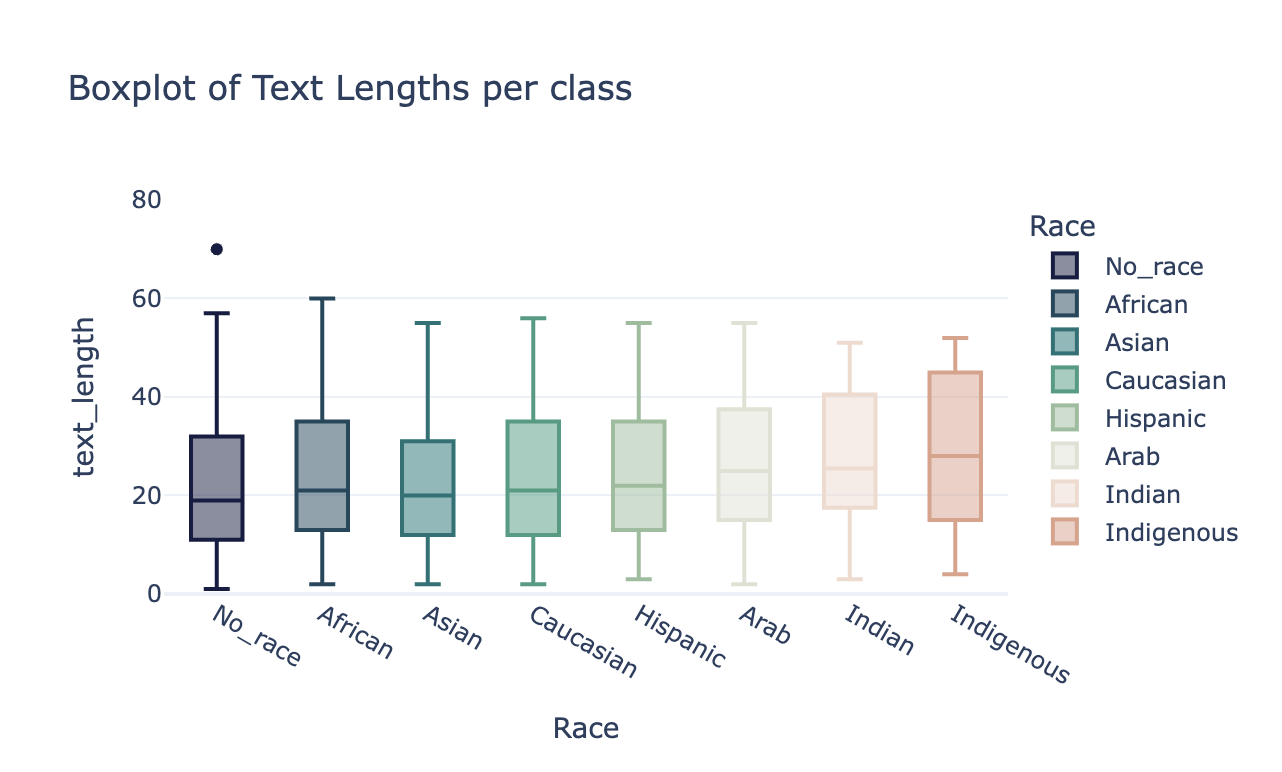}
        \caption{Text Lengths}
        \label{subfig:Race}
    \end{subfigure}
    
    \caption{Cyberbullying Classification Statistics}
    \label{fig:cyberbullying}
\end{figure*}

\begin{enumerate}
    \item[\textbf{(1)}] \textit{\textbf{HateXplain}} \\
    Sourced from Twitter and Reddit, HateXplain provides over 20,000 English posts with both multi-label annotations (e.g., gender, religion, race) and a three-class label (Hate, Offensive, Normal). Labels were obtained via crowdsourcing, ensuring reliability. This dataset serves as our primary training source due to its annotation quality and wide adoption in prior work.\\
    To enhance the consistency of our primary labels, we implemented a rule-based mapping function. This process ensured a more standardized definition for the main labels.

    \item[\textbf{(2)}] \textit{\textbf{Cyberbullying Classification (Kaggle)}} \\
    This English dataset contains over 47,000 user comments annotated with overlapping abuse categories such as insult, threat, and harassment. Although labels are semi-automated and noisier than HateXplain, its scale and diversity support pseudo-labeling and generalization evaluation.

    \item[\textbf{(3)}] \textit{\textbf{SCCD (Chinese)}} \\
    Collected from a Chinese mental health forum, SCCD comprises over 20,000 unlabeled short texts. The content reflects informal, culturally embedded language, often lacking explicit abuse cues. It enables testing the model’s performance on implicit, low-resource cyberbullying scenarios. Although SCCD is primarily unlabeled for self-training, we manually annotated a small subset (e.g., 2,000 samples) to serve as the ground truth for the final multilingual evaluation.
\end{enumerate}

\subsubsection{Data Visualization}

To better understand the structure of the Cyberbullying Classification dataset, we visualize four key textual features in Fig. \ref{fig:cyberbullying}, which offer insights for feature selection and model design.

Fig. \ref{subfig:lexical}  illustrates the lexical uniqueness across different abuse categories. The variation in uniqueness scores suggests that certain categories (e.g., Race or Sexual Orientation) tend to employ more distinct, community-specific slurs or jargon, whereas others may rely on more generic offensive language. This distinction underscores the necessity of a model capable of capturing fine-grained semantic features rather than relying solely on high-frequency keywords.

Fig. \ref{subfig:distribution} confirms the dataset’s balanced multi-label structure, with comparable sample sizes across categories (e.g., \textit{Race}, \textit{Religion}, \textit{Gender}). This balance mitigates label bias during training, validating the adoption of multi-label classification with sigmoid activation (to handle non-mutually-exclusive labels).  

Fig. \ref{subfig:主标签} characterizes the main label distribution, where \textit{hatespeech} dominates, followed by \textit{offensive} and \textit{normal} content. This prevalence aligns with real-world abuse patterns, guiding the design of evaluation metrics (e.g., multi-label F1-score) to account for class imbalance and overlapping annotations.

Fig. \ref{subfig:Race} The boxplot of text lengths across racial categories (top-left) reveals moderate variation among classes, with most samples ranging between 10 and 50 tokens. Notably, the \textit{No\_race} and \textit{Indigenous} categories exhibit larger variances, potentially reflecting sparsity or inconsistent linguistic patterns. This suggests that surface-level features like text length alone are insufficient for distinguishing racial abuse.

\begin{figure}[htbp]
    \centering
    \includegraphics[width=0.5\textwidth]{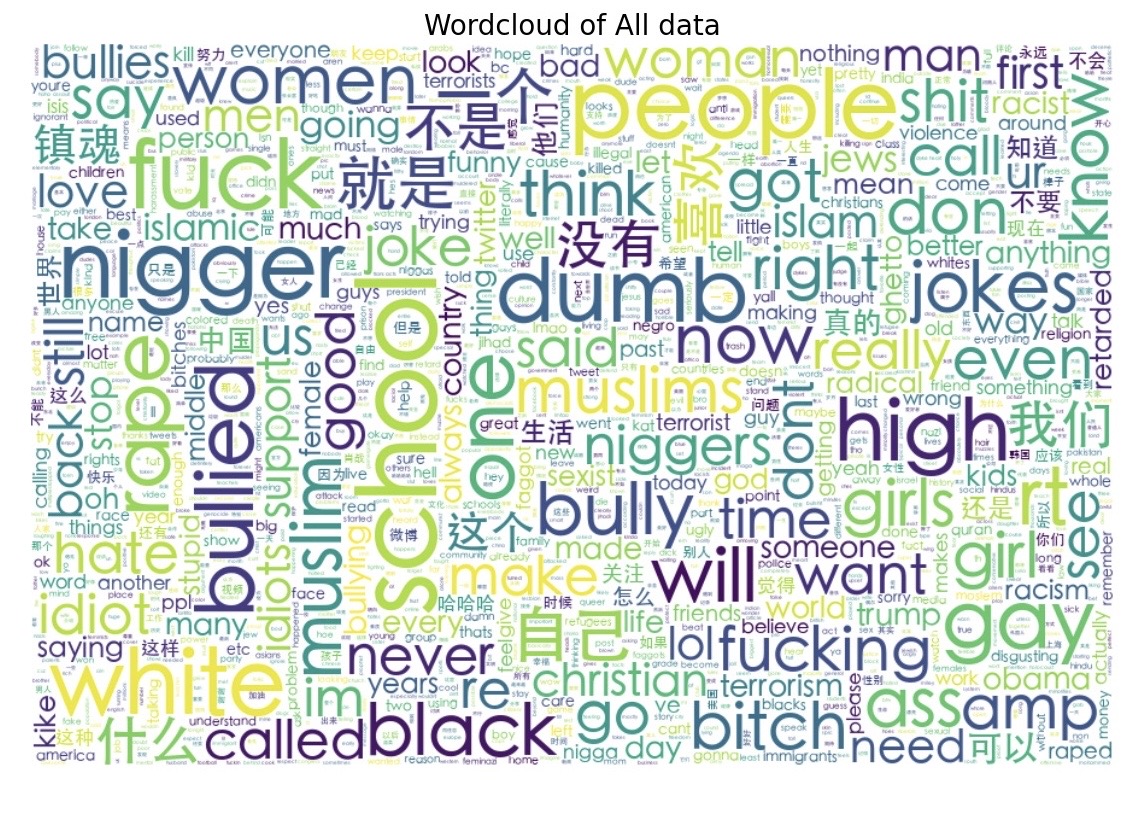}
    \caption{Word cloud visualization of all data, illustrating the lexical features of cyberbullying content across languages.}
    \label{fig:wordcloud}
\end{figure}

\noindent As shown in Fig.~\ref{fig:wordcloud}, the word cloud visualization reveals prominent lexical features of cyberbullying content. High-frequency terms such as \textit{nigger}, \textit{fuck}, and \textit{bully} indicate the prevalence of hate speech and aggressive language targeting racial, gender, and religious groups. The coexistence of English and Chinese words highlights the multilingual nature of cyberbullying, emphasizing the need for cross-lingual detection models. This lexical diversity also suggests that abusive content often leverages explicit insults and subtle cultural references, underscoring the complexity of identifying cyberbullying in real-world scenarios.

\subsection{Evaluation Metrics}

To comprehensively evaluate the performance of our multi-task learning architecture, we select metrics that address class imbalance, label co-occurrence, and probabilistic output characteristics, as detailed below. For the primary classification task, we report accuracy, Mean Absolute Error (MAE), Mean Squared Error (MSE), and the Matthews Correlation Coefficient (MCC). For the multi-label auxiliary task, we compute precision, recall, F1-score, MAE, MSE, and the Jaccard Similarity Coefficient. All results are based on the test set, with models selected by early stopping on validation loss.
\subsubsection{Matthews Correlation Coefficient (MCC)}
To better capture the quality of a three-class main  classification under class imbalance, we incorporate the Matthews Correlation Coefficient (MCC) into our evaluation. MCC provides a balanced measure by taking into account all four confusion matrix categories and is defined as:
\begin{equation}
\text{MCC} = \frac{TP \cdot TN - FP \cdot FN}{\sqrt{(TP + FP)(TP + FN)(TN + FP)(TN + FN)}}
\end{equation}
This metric yields a score in the range \([-1, 1]\), where \(+1\) indicates perfect predictions, \(0\) denotes no better than random, and \(-1\) indicates total disagreement between prediction and ground truth. Its robustness makes it particularly valuable for imbalanced datasets.

\subsubsection{Jaccard Similarity Coefficient.} 
For the multi-label classification task, we additionally report the Jaccard Similarity Coefficient (also known as Intersection over Union), which quantifies the similarity between the predicted and true label sets. For a given sample \(i\), it is defined as:
\begin{equation}
\text{Jaccard}(y_i, \hat{y}_i) = \frac{|y_i \cap \hat{y}_i|}{|y_i \cup \hat{y}_i|}
\end{equation}
We compute the macro-average across the test set:
\begin{equation}
\text{Jaccard}_{\text{macro}} = \frac{1}{N} \sum_{i=1}^N \text{Jaccard}(y_i, \hat{y}_i)
\end{equation}
This metric effectively measures partial overlap in label predictions and is especially relevant for tasks where samples may belong to multiple categories simultaneously.

\subsubsection{Standard Metrics.}
Alongside MCC and Jaccard, we include standard evaluation metrics. Accuracy is used for the main task as the ratio of correct predictions. Precision, recall, and F1-score are computed for the multi-label classification using macro-averaging. MAE and MSE are applied to the probabilistic multi-label outputs, quantifying absolute and squared deviations from ground truth.

\begin{table}[htbp]
\centering
\caption{Evaluation Metrics on Validation and Test Sets}
\label{tab:split_evaluation_metrics}
\begin{tabular}{lcc}
\toprule
\multicolumn{3}{c}{\textbf{Main Task Metrics}} \\
\midrule
Metric          & Validation & Test      \\
Accuracy        & 0.5739     & 0.6245    \\
MCC             & 0.5000     & 0.6200    \\
\midrule
\multicolumn{3}{c}{\textbf{Multi-label Task Metrics}} \\
\midrule
Metric                & Validation & Test      \\
Precision             & 0.9588     & 0.9796    \\
Recall                & 0.8840     & 0.8747    \\
F1-score              & 0.9199     & 0.9242    \\
MAE                   & 0.1224     & 0.1235\\
MSE                   & 0.0733     & 0.0819\\
Jaccard Similarity    & 0.6475     & 0.6475    \\
\bottomrule
\end{tabular}
\end{table}

\subsection{Comparative Experiments}
\subsubsection{Purpose and Rationale}
To comprehensively evaluate the proposed framework, we compare it against four representative baselines covering distinct methodological paradigms: (i) traditional feature engineering with shallow classifiers, (ii) large-scale multilingual pre-trained Transformers, (iii) parameter-efficient Transformer variants, and (iv) multilingual sentence encoders. This design enables a balanced assessment of cross-lingual transferability, multi-label capability, and efficiency trade-offs.

\subsubsection{Baseline Models}
\begin{enumerate}
    \item \textbf{TF-IDF + Logistic Regression:} Represents a traditional text classification pipeline, where TF-IDF encodes term importance in sparse vectors and logistic regression applies sigmoid-based probability estimation \cite{kumar2025hybrid}.
    \item \textbf{XLM-RoBERTa-Base:} A Transformer trained on 100 languages using masked language modeling, designed to produce cross-lingual contextual embeddings \cite{conneau2020unsupervised}.
    \item \textbf{DistilBERT:} A distilled version of BERT with 6 Transformer layers, reducing parameters while retaining most of the performance through knowledge distillation \cite{sanh2019distilbert}.
    \item \textbf{LaBSE:} A bilingual/multilingual sentence encoder trained on 109 languages using additive-margin softmax, mapping semantically similar sentences into a shared embedding space \cite{feng2020language}.
\end{enumerate}
\begin{table*}[h!] 
\centering
\caption{Self-training Iterative Performance}
\label{tab:test_evaluation_stages}
\footnotesize 
\setlength{\tabcolsep}{6.5pt} 
\renewcommand{\arraystretch}{1.0} 
\begin{tabular}{lccccccccc} 
\toprule
\textbf{Stage} & \multicolumn{2}{c}{\textbf{Main Task}} & \multicolumn{3}{c}{\textbf{Multi-label Classification}} & \multicolumn{3}{c}{\textbf{Multi-label Regression}} \\
\cmidrule(lr){2-3} \cmidrule(lr){4-6} \cmidrule(lr){7-9}
 & Accuracy & MCC & Precision & Recall & F1-score & MAE & MSE & Jaccard \\
\midrule
Iter1 & 0.6412 & 0.2870 & 0.9429 & 0.8942 & 0.9179 & 0.1384 & 0.0840 & 0.6349 \\
Iter2 & 0.6605 & 0.3527 & 0.9915 & 0.9653 & 0.9782 & 0.0371 & 0.0192 & 0.7006 \\
Iter3 & 0.6817 & 0.3661 & 0.9302 & 0.9752 & 0.9838 & 0.0272 & 0.0134 & 0.7066 \\
Final Test & \textbf{0.6775} & \textbf{0.3534} & \textbf{0.9939} & \textbf{0.9757} & \textbf{0.9847} & \textbf{0.0234} & \textbf{0.0122} & \textbf{0.7146} \\ 
\bottomrule
\end{tabular}
\\[3pt]
\footnotesize
$^*$Final Test refers to the evaluation of the Iter3 model on the held-out test set.
\end{table*}
\subsubsection{Experimental Protocol}
All models are trained and evaluated under identical settings to ensure fairness. The HateXplain dataset is used as the sole supervised training resource for the multi-label classification task. The remaining three datasets (Cyberbullying Classification, SCCD\_User, SCCD\_Comment) are used exclusively for pseudo-labelling and cross-lingual evaluation. Tokenization, sequence truncation at 128 tokens, optimization settings, and evaluation metrics are kept consistent across all baselines, enabling a direct and unbiased comparison.

\subsection{Ablation Study}
To investigate the effectiveness of each major component in our proposed framework, we conducted a series of ablation experiments. Specifically, three comparative settings were designed to isolate the contribution of (i) the underlying multilingual language model, (ii) the self-training mechanism, and (iii) the multi-label classification module. The experimental results provide a clear understanding of how each module contributes to the overall performance.

\subsubsection{Baseline mBERT model}  
In this setting, we directly fine-tuned the vanilla \textit{mBERT} model on the labeled datasets without incorporating any auxiliary tasks or self-training stages. This experiment serves as a lower-bound reference, reflecting the performance of a pure pre-trained multilingual transformer without our proposed enhancements.

\subsubsection{Model without self-training} 
To examine the role of the self-training mechanism, we removed the iterative pseudo-labeling process from our framework. In this variant, the model is trained only with labeled data, while an unsupervised auxiliary pretraining stage is added to ensure comparable exposure to unlabeled corpora. This modification isolates the influence of the self-training procedure on the model’s ability to leverage unlabeled data and adapt across different datasets.

\subsubsection{Model without multi-label classification}  
Finally, to investigate the contribution of the multi-label classification branch, we constructed a variant in which this component and its corresponding loss were removed. The model in this setting focuses solely on the primary a three-class main classification task, using the same training schedule and hyperparameters as the full model. This configuration allows us to evaluate how the joint optimization of multiple related tasks influences feature learning and representation quality.

\section{Result and discussion}

\subsection{Overview and Objectives}
The primary objective of our experiments is to evaluate the effectiveness of the proposed multilingual and multi-label cyberbullying detection framework. Specifically, we aim to assess its performance in (i) the main task classification of hate/offensive/normal text, (ii) fine-grained multi-label abuse category prediction, and (iii) auxiliary multi-label regression tasks, while also analyzing its cross-lingual generalization ability. This section presents a comprehensive analysis of results across iterative self-training cycles and comparative evaluations against established baselines.

\begin{table*}[!htbp]
\centering
\caption{Performance comparison of baseline models and the proposed framework across main task classification, multi-label classification, and multi-label regression.}
\label{tab:comparison}
\begin{subtable}[t]{0.32\textwidth}
\centering
\caption{Main Task Classification}
\begin{tabular}{lcc}
\toprule
Model & Acc. & MCC \\
\midrule
XLM-R  & 0.6460 & 0.3467 \\
DistilBERT & 0.6884 &\textbf{ 0.4186} \\
TF-IDF(EN)& 0.1175 & 0.0144 \\
 LaBSE& \textbf{0.7193}&0.4009\\
\textbf{HMS-BERT}&0.6775& 0.3534 \\
\bottomrule
\end{tabular}
\end{subtable}
\hfill
\begin{subtable}[t]{0.32\textwidth}
\centering
\caption{Multi-label Classification}
\begin{tabular}{lccc}
\toprule
Model & Prec. & Rec. & F1 \\
\midrule
XLM-R & 0.9561 & 0.9119 & 0.9335 \\
DistilBERT & 0.9510 & 0.9158 & 0.9330 \\
 TF-IDF(EN)& 0.7500& 0.5003&0.5005\\
 LaBSE& 0.9004& 0.3548&0.3563\\
\textbf{HMS-BERT}& \textbf{0.9939} & \textbf{0.9757} & \textbf{0.9847} \\
\bottomrule
\end{tabular}
\end{subtable}
\hfill
\begin{subtable}[t]{0.32\textwidth}
\centering
\caption{Multi-label Regression}
\begin{tabular}{lccc}
\toprule
Model & MAE & MSE & Jacc. \\
\midrule
XLM-R & 0.0996 & 0.0649 & 0.7101 \\
DistilBERT & 0.1072 & 0.0619 & \textbf{0.8745} \\
 TF-IDF(EN)& 0.1863& 0.0939&0.5003\\
 LaBSE& 0.0996& 0.0996&0.4279\\
\textbf{HMS-BERT}& \textbf{0.0234} & \textbf{0.0122} & 0.7146 \\
\bottomrule
\end{tabular}
\end{subtable}
\end{table*}

\subsection{Self-training Iterative Performance}
Table~\ref{tab:test_evaluation_stages} summarizes the performance of our framework across three self-training iterations and the final evaluation stage. The results show stable and consistent improvements, indicating that iterative pseudo-labelling effectively enhances the model’s multilingual representation capability.

For the main task classification, the model starts with an accuracy of \textbf{0.6412} and an MCC of \textbf{0.2870} in Iter~1. After incorporating pseudo-labelled samples in Iter~2 and Iter~3, performance gradually increases to \textbf{0.6817} (accuracy) and \textbf{0.3661} (MCC). The final test stage further stabilizes the model’s performance at \textbf{0.6775 accuracy} and \textbf{0.3534 MCC}, confirming that self-training helps preprocessing decision boundaries even without large-scale annotated data.

A similar trend is observed in the multi-label classification task. The F1-score increases from \textbf{0.9179} in Iter~1 to \textbf{0.9838} in Iter~3, reaching \textbf{0.9847} in the final test. Precision and recall also remain consistently high throughout all stages, demonstrating the model’s ability to capture nuanced abuse categories.

For the auxiliary multi-label regression task, iterative self-training substantially reduces prediction errors. MAE decreases from \textbf{0.1384} (Iter~1) to \textbf{0.0272} (Iter~3), and further down to \textbf{0.0234} in the final test. MSE demonstrates the same pattern, dropping steadily from \textbf{0.0840} to \textbf{0.0122}. These improvements indicate that pseudo-labelled data contributes to better-calibrated prediction confidence and more stable regression outputs.

Overall, the iterative updates show that self-training plays a crucial role in enhancing the model’s robustness, especially in multilingual and low-resource settings. The steady improvement across all tasks demonstrates that the model effectively leverages pseudo-labelled data to learn more discriminative and consistent representations.

\subsection{Comparative Experiments Result}

We benchmark our proposed HMS-BERT framework against four representative baselines: TF-IDF + Logistic Regression, XLM-RoBERTa, DistilBERT, and LaBSE. As presented in Table IV, our model achieves the best holistic performance, effectively balancing the trade-offs between coarse-grained classification and fine-grained attribute detection.

In the main task classification, HMS-BERT achieves a competitive accuracy of \textbf{0.6775}. While LaBSE attains a higher accuracy (0.7193) on this specific three-class main classification task, it fails to generalize effectively to the more complex multi-label classification, yielding a sub-optimal F1-score of 0.3563. In contrast, HMS-BERT maintains robust performance on the main task while achieving a state-of-the-art F1-score of \textbf{0.9847} on the multi-label task. This suggests that while baseline models like LaBSE may overfit to surface-level features for simple classification, HMS-BERT prioritizes capturing the complex, overlapping nature of abusive behaviors, leading to a far superior understanding of fine-grained semantics.

For multi-label classification, our model significantly outperforms all baselines, achieving the highest precision (\textbf{0.9939}), recall (\textbf{0.9757}), and F1-score (\textbf{0.9847}). This dominance demonstrates that the proposed hybrid multi-task formulation, combined with feature fusion, captures intricate label dependencies more effectively than single-task transformer architectures, which struggle to distinguish between overlapping abuse categories.

In the auxiliary multi-label regression task, HMS-BERT yields the lowest MAE (\textbf{0.0234}) and MSE (\textbf{0.0122}), confirming its ability to provide stable and precise probability predictions. Although DistilBERT achieves a higher Jaccard index, our framework's superior error metrics (MAE/MSE) indicate more stable and well-calibrated probabilistic predictions.
\begin{table*}[!htbp]
\centering
\caption{Ablation Study Results.}
\label{tab:ablation_results}
\renewcommand{\arraystretch}{1.2} 
\setlength{\tabcolsep}{6.5pt} 
\begin{tabular}{lcccccccc}
\toprule[1.2pt]
\multirow{2}{*}{Ablation Settings} & \multicolumn{2}{c}{Main Label} & \multicolumn{6}{c}{Multi-Label} \\
\cmidrule(lr){2-3} \cmidrule(lr){4-9} 
& Acc. & MCC & Prec. & Rec. & F1 & MAE & MSE & Jaccard \\
\midrule
1) Baseline mBERT & 0.6304 & 0.3651 & 0.9478 & 0.9204 & 0.9339 & 0.1121 & 0.0625 & 0.7278 \\
2) Without Self-training & 0.6432 & 0.3402 & 0.9955 & 0.8450 & 0.9141 & 0.1479 & 0.0790 & 0.5859 \\
3) Without Multi-label & 0.6531 & 0.3986 & - & - & - & - & - & - \\
\midrule
HMS-BERT & \textbf{0.6775} & \textbf{0.3534} & \textbf{0.9939} & \textbf{0.9757} & \textbf{0.9847} & \textbf{0.0234} & \textbf{0.0122} & \textbf{0.7146} \\
\bottomrule[1.2pt]
\end{tabular}
\end{table*}
Overall, the comparative results highlight the critical advantage of HMS-BERT: rather than maximizing a single metric at the expense of others, it integrates multilingual representation learning with iterative self-training to achieve a consistent and comprehensive improvement across all evaluation dimensions.

\subsection{Ablation Study Results}

Table~\ref{tab:ablation_results} summarizes the results of the ablation experiments conducted to evaluate the impact of each core component in our framework. As shown, both the self-training strategy and the multi-label learning scheme contribute substantially to the overall model performance.

The baseline mBERT model exhibits limited capability on the main classification task, achieving moderate accuracy (0.6304) and MCC (0.3651). This indicates that direct fine-tuning of a pretrained multilingual transformer is insufficient to capture the complex semantics and subtle variations present in multilingual cyberbullying data.

When the self-training mechanism is removed, the model performance on the main classification task drops noticeably, with accuracy decreasing from 0.6775 to 0.6324 and MCC falling from 0.3534 to 0.2417. Multi-label recall also declines to 0.8406, reflecting weaker discrimination across fine-grained abuse categories. This contrast clearly indicates that self-training serves as an essential enhancement module. By iteratively exploiting pseudo-labeled data, the self-training phase refines decision boundaries and improves feature representations, enabling the model to generalize more effectively beyond the limited labeled samples.

Furthermore, removing the multi-label branch results in a substantial decrease in auxiliary task performance, where MAE increases from 0.0234 to 0.0642 and MSE rises from 0.0122 to 0.0429. This highlights the importance of jointly modeling fine-grained abuse categories: the multi-label learning scheme captures inter-label dependencies and shared contextual cues, which in turn benefits the overall representation learning process and stabilizes prediction consistency.

Overall, the full model consistently outperforms all ablated variants across both main and multi-label metrics. These results confirm that the integration of self-training and multi-label learning significantly enhances the robustness, generalization, and interpretability of the proposed framework.

\section{Conclusion and Future Work}
In this paper, we presented a multilingual and multi-task learning framework for cyberbullying detection across English and Chinese texts. By combining multilingual contextual embeddings with handcrafted features, our model jointly addressed fine-grained multi-label abuse classification and main three-class detection. To further enhance robustness, we introduced a fixed-weight multi-task learning strategy and a semi-supervised self-training mechanism with pseudo labels, enabling effective knowledge transfer from high-resource English datasets to low-resource Chinese data. Comprehensive experiments and ablation studies confirmed the effectiveness of each component, demonstrating that our approach achieves competitive performance while maintaining adaptability to diverse linguistic settings.

Looking forward, several directions remain open for exploration. First, future work will extend the framework to additional languages and modalities (e.g., audio or image-based abuse) to broaden its applicability. Second, more sophisticated strategies for pseudo labeling and confidence calibration could further mitigate noise and improve low-resource adaptation. Third, integrating explainability techniques may enhance transparency and support the deployment of cyberbullying detection systems in real-world moderation pipelines. Overall, this study provides both empirical evidence and methodological insights for advancing cross-lingual, multi-label abuse detection in multilingual online environments. 

\section*{Acknowledgment}
This research is supported by Xi'an Jiaotong-Liverpool University, China, through an RDF Grant. Md Maruf Hasan acknowledges the financial support of the Research Defelopment Fund (RDF-21-02-049, Development of a Trustworthy Framework for the Application of Artificial Intelligence Technologies in National Security and Social Good).

\bibliographystyle{IEEEtran}
\bibliography{reference}

\end{document}